\documentclass[10pt,twocolumn,letterpaper]{article}

\usepackage{cvpr}              %
\usepackage[accsupp]{axessibility}
\usepackage[dvipsnames]{xcolor}

\definecolor{cvprblue}{rgb}{0.21,0.49,0.74}
\usepackage[pagebackref,breaklinks,colorlinks,citecolor=cvprblue]{hyperref}
\usepackage{epigraph} 
\usepackage[utf8]{inputenc}
\usepackage{csquotes}
\usepackage{wrapfig}
\usepackage{floatflt}
\usepackage[accsupp]{axessibility}

\usepackage[capitalize]{cleveref}
\usepackage{graphicx}
\usepackage{xspace}
\usepackage{booktabs}

\definecolor{MyDarkGreen}{rgb}{0.02,0.6,0.02}

\newcommand{\task}{temporal object intrinsics\xspace}

\newcommand{\Task}{Temporal Object Intrinsics\xspace}
\newcommand{\Pose}{Neural State Map\xspace}
\newcommand{\pose}{Neural State Map\xspace}
\newcommand{\poses}{Neural State Maps\xspace}
\newcommand{\posevol}{Neural Template\xspace}
\newcommand{\posevols}{Neural Templates\xspace}

\crefname{section}{Sec.}{Secs.}
\crefname{table}{Table}{Tables}
\crefname{figure}{Figure}{Figures}

\usepackage{multirow}
\usepackage{makecell}

\newcommand{\myparagraph}[1]{\vspace{0.1cm}\noindent\textbf{#1}}

\def\paperID{754} %
\def\confName{CVPR}
\def\confYear{2025}

\title{Birth and Death of a Rose}

\author{Chen Geng
\hspace{20pt}
Yunzhi Zhang
\hspace{20pt}
Shangzhe Wu
\hspace{20pt}
Jiajun Wu \\
Stanford University
}

\begin{document}

\twocolumn[
    \maketitle
    \begin{center}
    \vspace*{-15pt}
    \includegraphics[width=\textwidth]{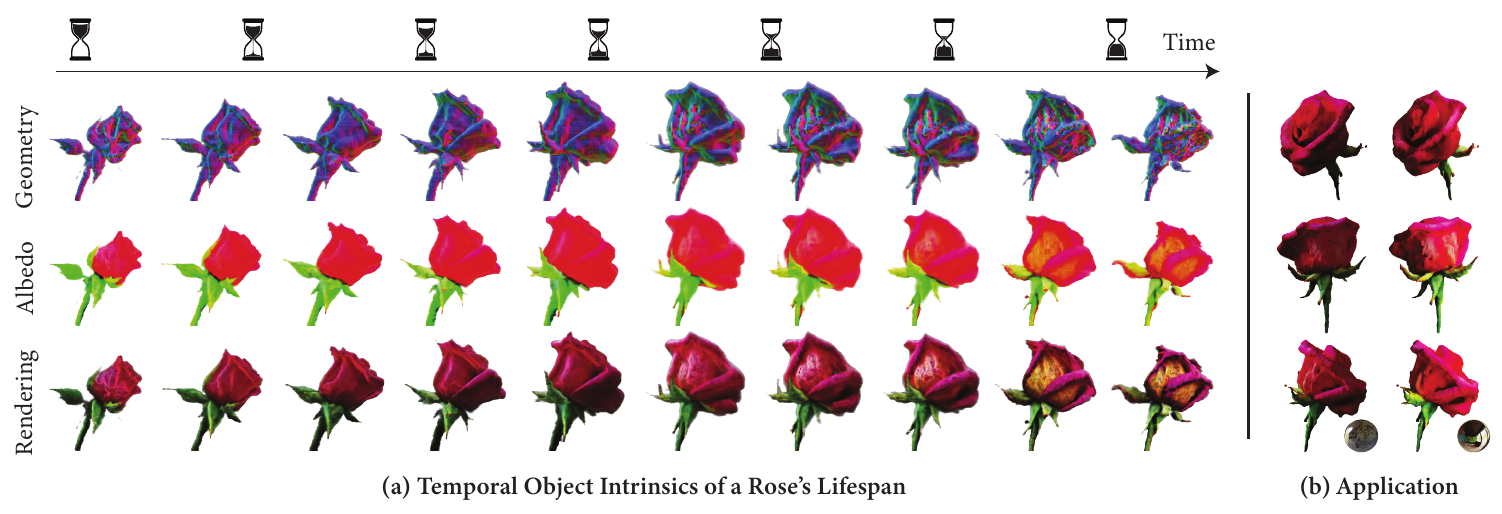}
    \vspace*{-20pt}
    \captionof{figure}{
    \textbf{The lifespan of a rose's \emph{object intrinsics} generated with our method.}
    We present a pipeline to generate temporally evolving sequences of 3D objects' geometry and material properties, including albedo, roughness, and metallic parameters (only albedo is shown here), as shown in (a), by distilling from 2D foundation models without any 3D data. As shown in (b), the generated assets can be rendered from any viewpoints and lighting conditions (environment map shown in the last row). 
    See the supplemental website for animations. 
    }
    \label{fig:teaser}
\end{center}
    \bigbreak
]
\begin{abstract}

\vspace{-5pt}
We study the problem of generating \task---temporally evolving sequences of object geometry, reflectance, and texture, such as a blooming rose---from pre-trained 2D foundation models. 
Unlike conventional 3D modeling and animation techniques that require extensive manual effort and expertise, we introduce a method that generates such assets with signals distilled from pre-trained 2D diffusion models.
To ensure the temporal consistency of object intrinsics, we propose \posevols for temporal-state-guided distillation, derived automatically from image features from self-supervised learning. 
Our method can generate high-quality \task for several natural phenomena and enable the sampling and controllable rendering of these dynamic objects from any viewpoint, under any environmental lighting conditions, at any time of their lifespan. Project website: {\url{https://chen-geng.com/rose4d}}.

\vspace{-10pt}
\end{abstract}    

\epigraph{Those granted the gift of seeing more deeply can see beyond form, and concentrate on the wondrous aspect hiding behind every form, which is called life.}{--- \textit{Hilma af Klint}}

\vspace{-10pt}
\section{Introduction}
\label{sec:intro}

As illustrated in \cref{fig:teaser}, starting from a bud, a rose gradually expands its petals and blossoms into full glory, only to wither and conclude its lifespan. This inevitable and unidirectional evolution—a journey shared by all living creatures on Earth—relentlessly and chronologically transforms their \textit{object intrinsics}: geometry, reflectance, and texture. 
Such progression of their object intrinsics characterizes our visual conception of the aging of an object, which we collectively refer to as its \emph{\task}.

Traditionally, creating graphics assets with realistic chronologically evolving object intrinsics requires extensive, object-specific manual effort and expertise \citep{przemyslaw1990algorithmic, Ijiri07Floral, Chaudhry2019ModellingAS}. Instead, we pursue a learning-based approach to generate such physically grounded graphics assets with no intervention. 
The generated \task can be seen as a ``3D time-lapse volumetric video'', from which we can sample instances and render them from any viewpoint, under any lighting condition, and at any time of their lifespan.

It is challenging to generate \task in a supervised manner due to the lack of annotated data. Therefore, we explore the potential of generative pipelines distilled from 2D diffusion models. 
Existing techniques like Score Distillation Sampling (SDS) \cite{poole2022dreamfusion} have shown promising results in 3D generation, but they are not directly transferable to our case of distilling \task with significant changes in their geometry and texture. In 3D distillation, it is already well known that SDS-like methods struggle to maintain 3D consistency due to the lack of 3D information in the 2D diffusion model, often referred to as the Janus problem~\cite{janus}. Unfortunately, the situation worsens in our optimization of 4D representations due to global inconsistency across both space and time: not only may a signature view appear from multiple camera viewpoints, but a common temporal state may also appear repeatedly throughout the entire duration. 

To mitigate such 4D inconsistencies, we propose \emph{\posevols} for temporal-state-conditioned distillation.
\posevol, a mapping that takes in viewpoint and time and outputs the ``temporal state'' information of modeled natural process, captures the lifespan of dynamic object's intrinsics; it can be automatically constructed by forging self-supervised image features \cite{oquab2023dinov2,caron2021emerging} obtained from a rough initial 4D reconstruction depicting the dynamic process. It allows us to anchor the distillation gradients to a particular viewpoint and timestamp by conditioning the diffusion model on a 2D \pose representing the temporal state. This significantly improves the distillation efficiency and 4D consistency, as each view receives distillation signals tailored to that particular temporal state.

To model the photorealistic texture of an object, we further decompose its appearance into physically-based surface materials components and recover these representations during the distillation process using a differentiable PBR renderer. We also propose a hybrid 4D representation for consistent yet high-fidelity generations.

To summarize, we propose a new task of generating \emph{\task}, in the form of temporally-evolving sequences of 3D shape, reflectance, and texture.
We introduce a framework to distill 4D-consistent \task from pre-trained 2D diffusion models.
Central to this framework is a canonical \posevol that anchors the distillation signals to specific temporal states.
We test this framework on several different object categories. We quantitatively compare its performance with prior state of the art on automatic 4D generation, suggesting advantages of the proposed methods across different examples. A further ablation study shows that the core modules and techniques proposed in this framework are crucial for performance. 

\begin{figure*}[t]
  \centering
  \includegraphics[width=0.9\linewidth]{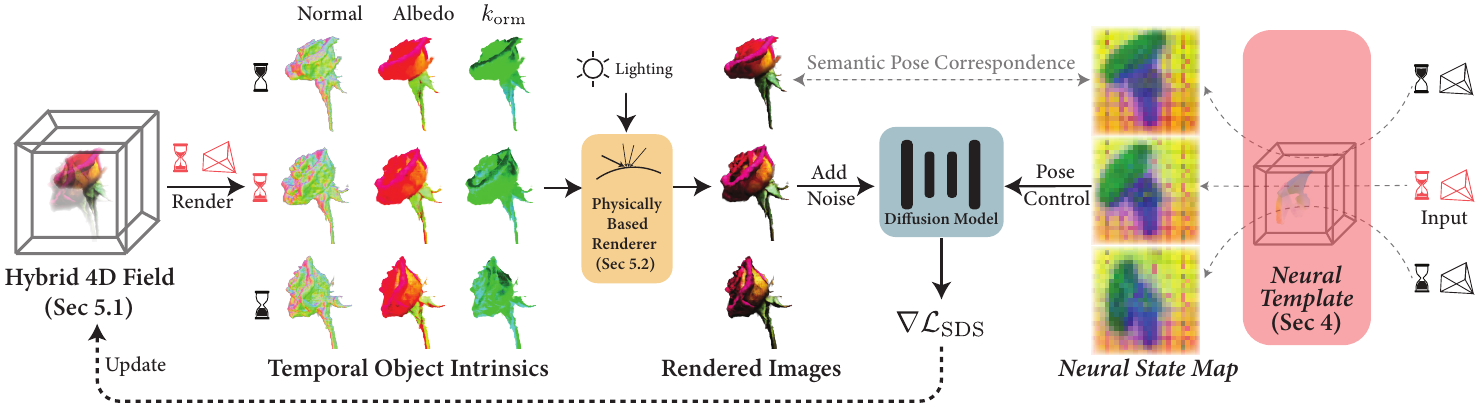}
  \vspace{-0.3cm}
  \caption{ \textbf{Overview of the process to generate \task.} For a modeled natural process, we first estimate a task-specific \emph{\posevol} to get a 4D-consistent representation of its temporal state changes~(\cref{sec:method_posevol}). Then we iteratively optimize a hybrid 4D intrinsics field (\cref{ssec:hybrid}) to generate \task. In each iteration of optimization, we sample a camera pose and timestamp to render this optimizable 4D field into intrinsic maps. These maps are subsequently rendered into RGB images using a physically-based renderer~(\cref{ssec:shading}). Concurrently, the same camera pose and timestamp are used to query the \posevol for the corresponding \pose. This map conditions a fine-tuned 2D diffusion model, which provides guidance signals to update the 4D representation. See \cref{sec:method_posevol} and \cref{fig:nt} for details on \posevol. 
  }
  
  \vspace{-8pt}
  \label{fig:overview}
\end{figure*}

\begin{figure}[t]
  \centering
  \includegraphics[width=\linewidth]{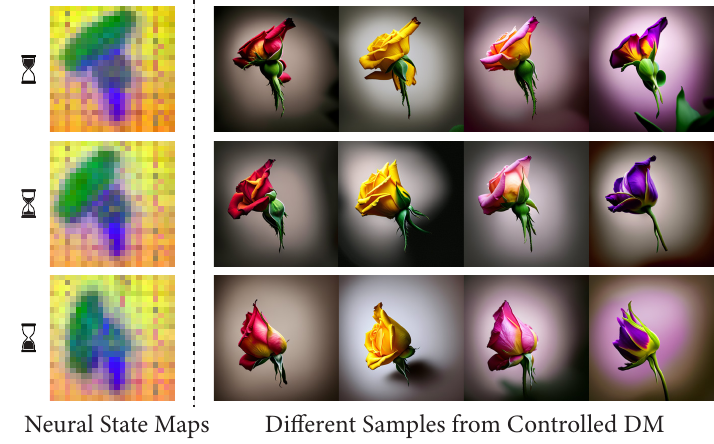}
  \caption{\textbf{A motivating example: \emph{\pose} as a conditional control signal for diffusion models.} Compressed DINOv2 features, or ``\pose'', represent both the temporal state and spatial viewpoint information throughout a temporal natural process, and inject effective 4D control into 2D diffusion models. }
  \vspace{-10pt}
  \label{fig:pseudo-pose}
\end{figure}

\section{Related Work}
\label{sec:related}

\myparagraph{Dynamic Modeling and Generation.}
Understanding and further synthesizing the dynamic motion of objects have been a challenging and long-standing task in Computer Graphics.
A prominent challenge is the limited amount of data due to the high cost of motion data collection. Prior works address this challenge via simulation-based techniques~\cite{przemyslaw1990algorithmic, Ijiri07Floral, Chaudhry2019ModellingAS} or 
human-designed templates~\cite{petrovich2021actionconditioned,loper2023smpl}, but the assumption of access to the simulator or template limits applying such methods to generic object categories. More recently, several methods have weakened the template assumption and required only the skeleton structure by introducing skinning-based representation, learning from pure image~\cite{wu2023magicpony,li2024fauna,yao2022lassie,yao2023hilassie} or video data~\cite{yang2022banmo} using pixel reconstruction as objectives, and are extended to the task of 4D generation~\cite{sun2023ponymation}. These methods have been mostly applied to articulated objects or animals, while in comparison, our method can be applied to generic objects such as flowers, which require a highly expressive representation such as the proposed \posevol.

\myparagraph{Controlling and Distilling from Diffusion Models.}
One of the applications of modeling 4D intrinsics is to synthesize 4D videos of an object similar to the input, which can be cast into a controllable generation problem. Recent methods leverage pre-trained image~\cite{rombach2021highresolution} or video diffusion models~\cite{blattmann2023stable} to support conditioning signals such as camera viewpoints~\cite{liu2023zero}, lighting conditions~\cite{kocsis2024lightit,deng2024flashtex,Jin2024NeuralGR,Zeng2024DiLightNetFL}, type of motions~\cite{zhao2023motiondirector}, and identity personalization~\cite{ruiz2022dreambooth,gal2022textual}. These methods typically train an additional lightweight network~\cite{zhang2023adding}, low-rank layers~\cite{hu2021lora}, or explicitly distill the pre-trained model into a new representation~\cite{poole2022dreamfusion,metzer2022latent,dong2024coin3d,consistent4d}. Ours falls into the last category; compared with previous work that uses score distillation to generate 4D contents~\cite{4dfy,consistent4d,dreamgaussian4d,wang2024vidu4d,sun2024eg4d,yu20244real,ren2024l4gm,alignyourgaussian,gao2024gaussianflow,xie2024sv4d,diffusion2,li2024dreammesh4d,yuan20244dynamic,uzolas2024motiondreamer,liang2024diffusion4d}, our method supports the modeling of lighting, viewpoint, and timestep all at once and tackles the full task of 4D intrinsic modeling, while prior works tackle parts of the task. 

\myparagraph{Intrinsic Decomposition.}
Recovering object intrinsics, namely its geometry, texture, and material, is a long-standing task~\cite{barron2014shape}. This information is not directly available in image or video observations, and previous methods tackle this task by optimizing for pixel reconstruction~\cite{barron2012color,boss2021nerd,physg2021,zhang2021nerfactor,jin2023tensoir}, learning with supervised learning targets using ground truth material from synthetic data~\cite{lee2024dmp}, or generative modeling~\cite{zhang2023seeing,violante2024physically,fantasia3d,zhang2024dreammat,richardson2023texture,xu2023matlaber,deng2024flashtex,Zeng2024DiLightNetFL,Jin2024NeuralGR}. However, these methods exclusively focus on static intrinsics, while our method, in addition, accounts for the temporal evolutionary nature of these properties. 

\begin{figure*}[t]
  \centering
  \includegraphics[width=\linewidth]{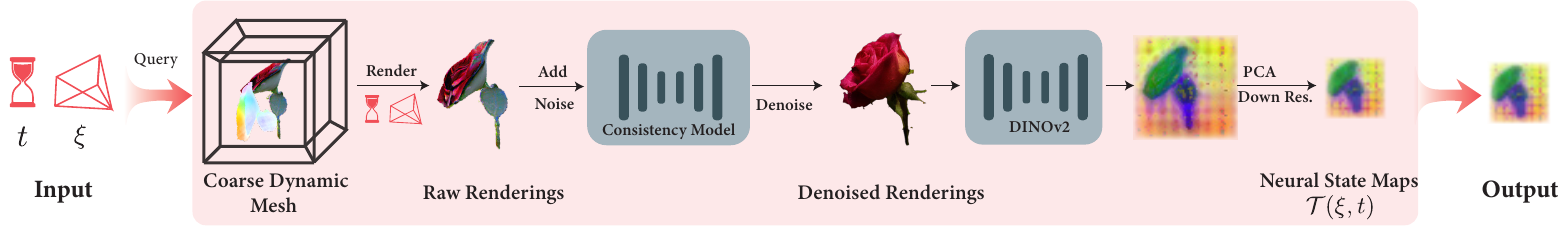}
  \vspace{-0.4cm}
  \caption{\textbf{Internal structure of \posevol.} Taking camera pose $\xi$ and timestamp $t$ as input, the neural template first render a task-specific coarse dynamic mesh under the provided condition. It is further denoised with a consistency model and encoded to a feature map, which will be further used to control the diffusion models.}
  \vspace{-15pt}
  \label{fig:nt}
\end{figure*}
\section{Overview}
\label{sec:overview}

\subsection{Task}

Our goal is to generate \emph{\task}, a chronologically ordered sequence of the \emph{object intrinsics} in a natural process, such as ``\textit{a rose blooming}'', where ``object intrinsics'' refers to the intrinsic properties of an object---in our case, its geometry and material.

The natural processes considered in this paper are temporally evolving transformations imposed on the object by the external environment. Formally, these processes are characterized by several key properties: they span a long-range time frame, from hours to days; they are inevitable, driven by external factors; they involve significant changes in the object's geometry and/or material throughout its lifespan; and they occur in chronological sequence with a unidirectional order of temporal state changes. 

\subsection{Method Motivation}
\label{subsec:motivation}

We propose to distill 4D \task from 2D foundation models. Recent works have utilized techniques such as SDS~\cite{poole2022dreamfusion,sjc} to generate 4D content~\cite{dreamgaussian4d, 4dfy}. However, these methods practically do not perform well on this task. On the one hand, they model appearance with view-dependent radiance, which may not be fully physically accurate and does not support relighting. More critically, as shown in our experiments (\cref{fig:comp4d}), those methods typically generate minor and unrealistic motions, insufficient for representing the significant temporal state changes observed in objects undergoing drastic transformations over their lifespan.

For prior works, the primary challenge arises from the discrepancy between the spatial capacity of 2D diffusion models and the 4D information needed in the \task. It is well-known that common score-distillation-based 3D generation methods~\cite{poole2022dreamfusion,sjc} often result in ``Janus problem'', where a signature view repeats itself in various sides of the generated asset.
This is mainly because the 2D diffusion model has limited knowledge of the 3D viewpoint control. In the case of 4D generation, apart from repetitive views, the generated instances may exhibit repetitive \textit{temporal states} across different timestamps, due to the lack of sufficient temporal anchoring. 

Therefore, distilling from 2D diffusion models necessitates a clear 3D and temporal control signal~\cite{zhang2023adding} to enable high-fidelity 4D object intrinsics generation. What would be a good control signal? In traditional Computer Graphics literature~\cite{kavan2005spherical,kavan2007skinning,loper2023smpl}, \textit{skeletons} are commonly used to represent the motion state. As we are interested in generic objects beyond articulated ones, we need a more generic representation that shares a similar role as skeletons to encode semantic affinity across object parts. 

In fact, after spatial downsampling and Principal Component Analysis\xspace(PCA), 2D image feature maps extracted using DINOv2~\cite{oquab2023dinov2}
can effectively provide such semantic affinity information,
which is similar to the projection of skeleton in traditional animation. As shown in \cref{fig:pseudo-pose}, these feature maps effectively encode the temporal state as well as the viewpoint information for a natural process. We therefore call those maps ``\poses''. We empirically found that these maps can serve as conditional signals for ControlNet~\cite{zhang2023adding} to control the temporal state and viewpoint of generated images. 

This finding bridges the 2D diffusion model and 4D information naturally. For a natural process, we can construct a canonical representation that tells us the temporal state information given a query camera viewpoint $\xi$ and a timestamp $t$. We call this a ``\posevol'', which defines a mapping $\mathcal{T}(\xi, t)$ from the camera pose and timestamp to the \pose. We will further describe how we construct such mappings in \cref{sec:method_posevol}.

\subsection{System Overview}

We build a system that takes in the prompt of a natural process, such as ``\textit{a rose blooming}'', and generate \task, as shown in \cref{fig:overview}. The generation process is split into two stages. In the first stage, we construct a coarse dynamic mesh to represent the 4D temporal stages in the natural process. The constructed coarse dynamic mesh can be used to obtain the \posevol as in \cref{fig:nt}, which will be used in the second stage. We discuss the details of this process in \cref{sec:method_posevol}.

In the second stage, we generate \task by iteratively optimizing a hybrid 4D object intrinsics representation with gradients from 2D diffusion models. At each iteration, we randomly sample a camera viewpoint and a timestamp, and render an image with a physically-based renderer. The rendered image is then perturbed with noise and fed into a diffusion model conditioned on the \pose, to obtain a generative score for updating the hybrid 4D field. This process is further discussed in \cref{sec:method_synthesis}.

\section{\posevol}
\label{sec:method_posevol}

As introduced in \cref{sec:overview}, the \posevol aims to capture the canonical temporal state changes during the modeled process. Formally speaking, it is a mapping that takes a query viewpoint $\xi\in\mathbb{SE}(3)$ and a timestamp $t\in\{1, \cdots, T\}$, where $T$ is the total number of timestamps, and outputs a corresponding ``\pose'' $\mathcal{T}(\xi, t)\in\mathbb{R}^{H_F \times W_F \times d_F}$ (recall that it is a 2D feature map from DINOv2 after PCA and down-resolution).

The internal structure of \posevol can be found in \cref{fig:nt}. It consists of a task-specific coarse dynamic mesh, which stores coarse canonical pose information of the progressive state changes throughout the natural process, and several pre-trained frozen models that convert the renderings from the canonical RGB field to the final \pose. We first discuss how we obtain the task-specific coarse dynamic mesh (\cref{ssec:method_distillation}), and then illustrate the algorithm to obtain the output \pose (\cref{ssec:feat_render}). More details of this section can be found at the \text{Sec.\,A} of the supplementary material.

\subsection{Estimating Coarse 4D Geometry} 
\label{ssec:method_distillation} 

For each natural process considered, we construct a coarse dynamic mesh to represent the semantic structure throughout the process. Such coarse dynamic mesh does not need to produce high rendering quality. Rather, it only needs to represent the coarse shape and appearance changes throughout the process. 

We present one way to obtain such a dynamic mesh. Assuming that we have a 2D video diffusion model capable of modeling the motion of the natural process\footnote{Practically, we train a LoRA~\cite{hu2021lora} on a set of manually-collected time-lapse videos on pre-trained video diffusion models~\cite{yang2024cogvideox}. See supplementary material for details.}, we can sample a video $V_\text{ref}$ from this model. We perform a physically-grounded 4D reconstruction of the video $V_\text{ref}$ to obtain the dynamic mesh, using the following procedure.

We first choose one frame in the video as the canonical frame, and use off-the-shelf 3D reconstruction models~\cite{liu2023zero,wang2023imagedream} 
to obtain a static 3D mesh as the representation of the static canonical frame.
We then optimize a deformation field $D(\xi,t)$ that explicitly deform the 3D canonical mesh to match the geometry and appearance of other frames in the video.
During each iteration, we rasterize the 3D scene flow into 2D flow maps and supervise them with optical flow estimated from off-the-shelf models~\cite{shi2023videoflow}. We use additional regularization terms including As-Rigid-As-Possible (ARAP)~\cite{sorkine2007rigid,wu2023dove,igarashi2005rigid} to ensure the physical plausibility. More details can be found in the Sec. A of the supplementary material.

\subsection{Extracting \Pose}
\label{ssec:feat_render}

Given a camera pose $\xi$ and a timestamp $t$, we render the aforementioned dynamic mesh into an RGB image. In order to use this image to control the diffusion model, we map it to a \pose defined in \cref{subsec:motivation}, as described next.

\myparagraph{Denoising raw renderings with consistency model.} Existing 3D/4D reconstruction models typically have difficulties in accurately reconstructing rarely-seen natural objects, such as \textit{roses}. Therefore, the coarse dynamic mesh obtained above might not align with the natural image distribution, on which the foundation model DINOv2 is trained.

To address this issue, we propose to use consistency distillation to aggregate priors from 2D generative models while adhering to the \pose features. 
For any given pre-trained diffusion model, we use a consistency model~\cite{luo2023latent,song2023consistency} distilled from it, denoted as $\hat\epsilon_\gamma$, to obtain the denoised samples of the noisy renderings. 
Let
\begin{equation}
    \mathbf{f}_\gamma(\mathbf{z}, \mathbf{c}, \tau) = c_\text{skip} (\tau) \cdot \mathbf{z} + c_\text{out} (\tau) \left(\frac{\mathbf{z} - \sigma_\tau \hat\epsilon_\gamma \left(\mathbf{z}, \mathbf{c}, \tau\right)}{\alpha_\tau}\right),
\end{equation}
where $\tau$ is the diffusion step, $c_\text{out}$ and $c_\text{skip}$ are functions satisfying certain boundary conditions \cite{song2023consistency}, $\sigma_\tau,\alpha_\tau$ follow a diffusion noise schedule, $\mathbf{z}$ is the latent code with noise added at noise level $\tau$, and $\mathbf{c}$ is the text prompt corresponding to the input video, \eg, `rose'.

\myparagraph{Encoding RGB map into \pose.} After obtaining denoised image with this process, we use a pre-trained 2D self-supervised image encoder $F$~\cite{oquab2023dinov2} to obtain a feature map, which is further downsampled after a PCA step to encourage the \pose to capture only low-frequency information.

\begin{figure*}[t]
  \centering
  \includegraphics[width=0.85\linewidth]{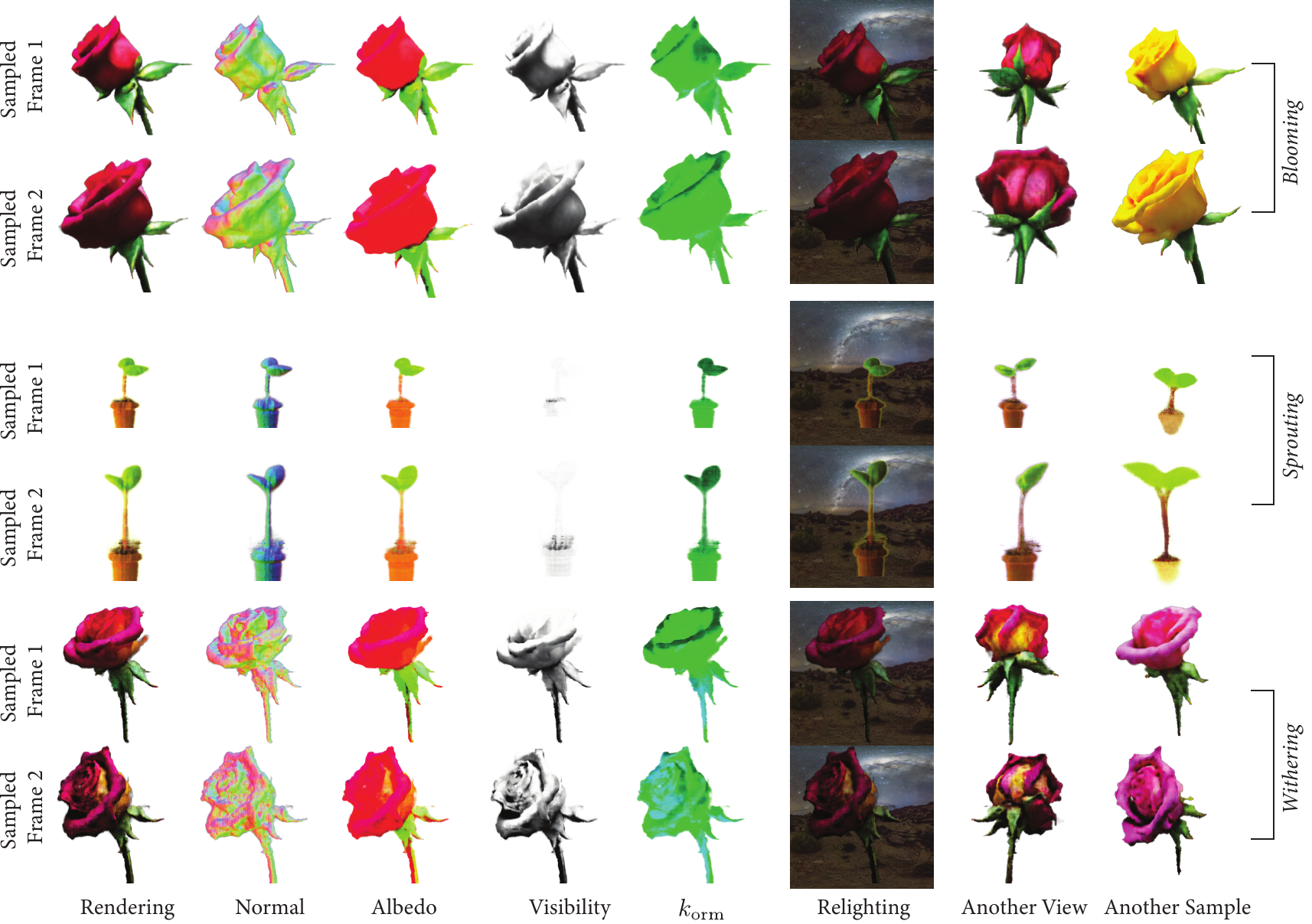}
  \vspace{-0.2cm}
  \caption{\textbf{Results on \task generation.} We show the generated 4D sequence with our method on part of the evaluated phenomena. Our method synthesizes the geometry and albedo, as well as material properties of roughness and metallic, with high fidelity. The 4D sequence can be used for relighting and novel view synthesis. We also show another sample in the generation. Please check the supplemental material for more results.
  }
  \vspace{-10pt}
  \label{fig:major}
\end{figure*}

\section{Guided 4D Object Intrinsics Synthesis}
\label{sec:method_synthesis}

The estimated \posevol $\mathcal{T}$ from \cref{sec:method_posevol} can be used as guidance for generating a 4D instance $\mathcal{G}$. 
We first describe our 4D representation for $\mathcal{G}$ (\cref{ssec:hybrid}) and its rendering process (\cref{ssec:shading}), and then explain how to distill $\mathcal{G}$ via 2D diffusion models controlled with $\mathcal{T}$ (\cref{ssec:controlnet}).

\subsection{Hybrid 4D Representation}
\label{ssec:hybrid}
We now introduce an implicit neural 4D representation for $\mathcal{G}$ that is both expressive and temporally consistent. The geometry of $\mathcal{G}$ is modeled as a time-evolving implicit level-set~\cite{wang2023neus2,mehta2022level,Novello_2023_ICCV}, represented by an implicit neural SDF. Given a spatial coordinate $\mathbf{x}$ and timestamp $t$, we compute a feature vector $f_{\text{4D}}(\mathbf{x}, t)$, which is subsequently passed through two MLPs. These MLPs output SDF values and material parameters essential for neural rendering~\cite{wang2021neus,mildenhall2020nerf} as detailed in \cref{ssec:shading}.

The neural 4D feature vector $f_{\text{4D}}$ is computed via a hybrid 4D representation with K-Planes~\cite{kplanes2023} representing low-frequency information and Neural Graphical Primitives (NGP)~\cite{muller2022instant} representing high-frequency details.

For K-Planes, we maintain $6$ planes denoted with $\mathbf{P}_c$ to capture temporal information with tensorial decomposition. The low-frequency feature of K-Planes $f_{\text{low}}$ is calculated with
\begin{equation}
    f_{\text{low}}(\mathbf{x}, t) = \prod_{c \in C} \psi (\mathbf{P}_{c}, \pi_c(\mathbf{x}, t)),
\end{equation}
where $\pi_c$ projects $(\mathbf{x}, t)$ onto the $c$'th plane, $\psi$ denotes interpolation, and $\Pi$ denotes the Hadamard product. 

We find that $f_\text{low}$ represents temporal consistency information but struggles to represent high-frequency details. To address this, we represent the high-frequency details using several NGPs for $K$ keyframes. For each of the keyframes, a multi-resolution hash grid is constructed. The feature for a rendered timestamp $t$ is further obtained with
\begin{equation}
    f_{\text{high}} (\mathbf{x}, t) = \textbf{lerp}\left(\frac{t - t_i}{t_{i+1} - t_i},  \psi_{t_i}(\mathbf{x}), \psi_{t_{i+1}}(\mathbf{x})\right), t_i \leq t \leq t_{i+1},
\end{equation}
where $\psi_t$ is the NGP encoding for timestamp $t$, and $t_i$ and $t_{i+1}$ are two keyframes around the sampled timestamp. The full 4D feature $f_\text{4D}$ is computed by concatenating $f_\text{low}$ and $f_\text{high}$. 

\subsection{Physically-Based Rendering}
\label{ssec:shading}
We now describe the rendering procedure for the hybrid representation of $\mathcal{G}$ from \cref{ssec:hybrid}. 
\paragraph{Shading.} Following common conventions, we use the Disney Principled PBR material model~\cite{Burley2012PhysicallyBasedSA} to perform the shading process. 
Specifically, the material parameters consist of diffuse color $k_d$, the visibility map $V(\mathbf{x})$, and $k_\text{orm} = (o, r, m)$ with roughness $r$, metallic parameter $m$, and $o$ left unused. 
The specular color $k_s$ is computed with $k_s = (1 - m) \cdot 0.04 + m \cdot k_d$. The outgoing radiance $L(\omega_o)$ in direction $\omega_o$ can be represented with the rendering equation \cite{kajiya1986rendering}:
\begin{equation}
L(\mathbf{x}, \mathbf{\omega}_o) = \int_\mathbf{\omega} L_i(\mathbf{x}, \mathbf{\omega}_i) f(\mathbf{\omega}_i, \mathbf{\omega}_o) V(\mathbf{x}) (\mathbf{\omega}_i \cdot \mathbf{n}) \mathrm{d} \mathbf{\omega}_i, 
\end{equation}
where $\mathbf{n}$ is the surface normal, $L_i(\mathbf{x}, \omega_i)$ is the incoming radiance from direction $\omega_i$, $f(\omega_i, \omega_o)$ is the BSDF computed from the material parameters. 
\paragraph{Volume rendering.} After computing the shading value at each spatial coordinate $\mathbf{x}$ and timestamp $t$, we follow a standard differentiable rendering process~\cite{mildenhall2020nerf,wang2021neus} to produce the rendered frame at time $t$. These frames are then concatenated to form a complete video $V_G$. More details on rendering can be found in the supplementary material.

\subsection{4D Distillation with \pose Controls} To obtain parameters $\theta$ of 4D content $\mathcal{G}$ using the above representation, we propose an optimization framework that distills from pre-trained 2D diffusion models~\cite{rombach2021highresolution}, by adapting these models to condition on \pose computed from $\mathcal{T}$ as discussed in \cref{sec:overview}. 
\label{ssec:controlnet}

Specifically, to inject \pose conditions, we follow ControlNet~\cite{zhang2023adding} and train backbone models on synthetic data. 
The training dataset consists of data pairs $(I, F(I))$, where $I$ is an image generated from open-accessible APIs~\cite{imagen} and $F(I)$ is the control signal computed with neural pose encoder $F$ from \cref{sec:overview}. 
The trained models are used to guide optimization of $\mathcal{G}$.

We optimize the representation in \cref{ssec:hybrid} with score distillation from \pose-controlled diffusion models (\cref{ssec:controlnet}). As each optimization iteration, we sample a camera viewpoint $\xi$ and a timestamp $t$, and compute the following SDS gradient:
\begin{equation}
\begin{aligned}
    &\nabla_\theta \mathcal{L}_\text{SDS} (\theta; \xi, t, \mathbf{y} ) = \\
    &\mathbb{E}_{\tau, \epsilon, \xi, t} \left[
        w(\tau) \left( \hat\epsilon \left( \mathbf{z}_\tau; \tau, \mathcal{T}(\xi, t), \mathbf{y} \right) - \epsilon \right)
        \frac{\partial \mathbf{z}_\tau}{\partial \theta}
    \right],
\end{aligned}
    \label{eq:imgsds}
\end{equation}
where $w(\tau)$ is a weighting function, $\mathbf{y}$ is the text prompt, $\mathbf{z}_\tau$ is the latent code of image rendered with the physically-based renderer, with a noise level $\tau$. 

We also include a temporal regularization term to ensure the temporal consistency of the generated $\mathcal{G}$. We render a video $V_\mathcal{G}(\xi)$ with viewpoint $\xi$. It is then fed into a video diffusion model~\cite{yang2024cogvideox} with added noise and is denoised into a refined video $\hat{V}_\mathcal{G} (\xi)$ with a better temporal consistency. The refined version is used to supervise the representation with $\mathcal{L}_\text{Vid} (\theta; \xi) = ||V_\mathcal{G} (\xi) - \hat V_\mathcal{G}(\xi)||_2^2$. More details on the optimization can be found in the \text{Sec.\,B}.

\begin{table}
\centering
\resizebox{\linewidth}{!}{%
\begin{tabular}{@{}lllll@{}}
\toprule
                & \multicolumn{3}{c}{User Study}                         & \multicolumn{1}{c}{\multirow{2}{*}{CLIP $\uparrow$}} \\ \cmidrule(r){2-4}
                & Motion Alignment $\uparrow$ & Motion Realism $\uparrow$  & Visual Quality  $\uparrow$ & \multicolumn{1}{c}{}                      \\ \midrule
DG4D~\cite{dreamgaussian4d} & 2.26\%           & 5.26\%           & 11.28\%          & 29.65                                     \\
4D-Fy~\cite{4dfy}           & 4.21\%           & 11.13\%          & 21.95\%          & 30.11                                     \\
STAG4D~\cite{stag4d}          & 16.69\%          & 13.83\%          & 5.71\%           & 28.01                                     \\
\midrule
Ours            & \textbf{76.84\%} & \textbf{69.77\%} & \textbf{61.05\%} & \textbf{30.83}                            \\ \bottomrule
\end{tabular}
}
\caption{\textbf{Comparison with 4D generation baselines.} We perform a user study on 7 different natural processes to assess the performance of our method and previous 4D generation methods in three aspects. 
We report the percentage of cases in which a method ranks first, which indicates a significant advantage of our method. We also report a CLIP score. }
  \vspace{-15pt}
\label{tab:user-study}
\end{table}

\begin{figure*}[t]
  \centering
  \includegraphics[width=0.9\linewidth]{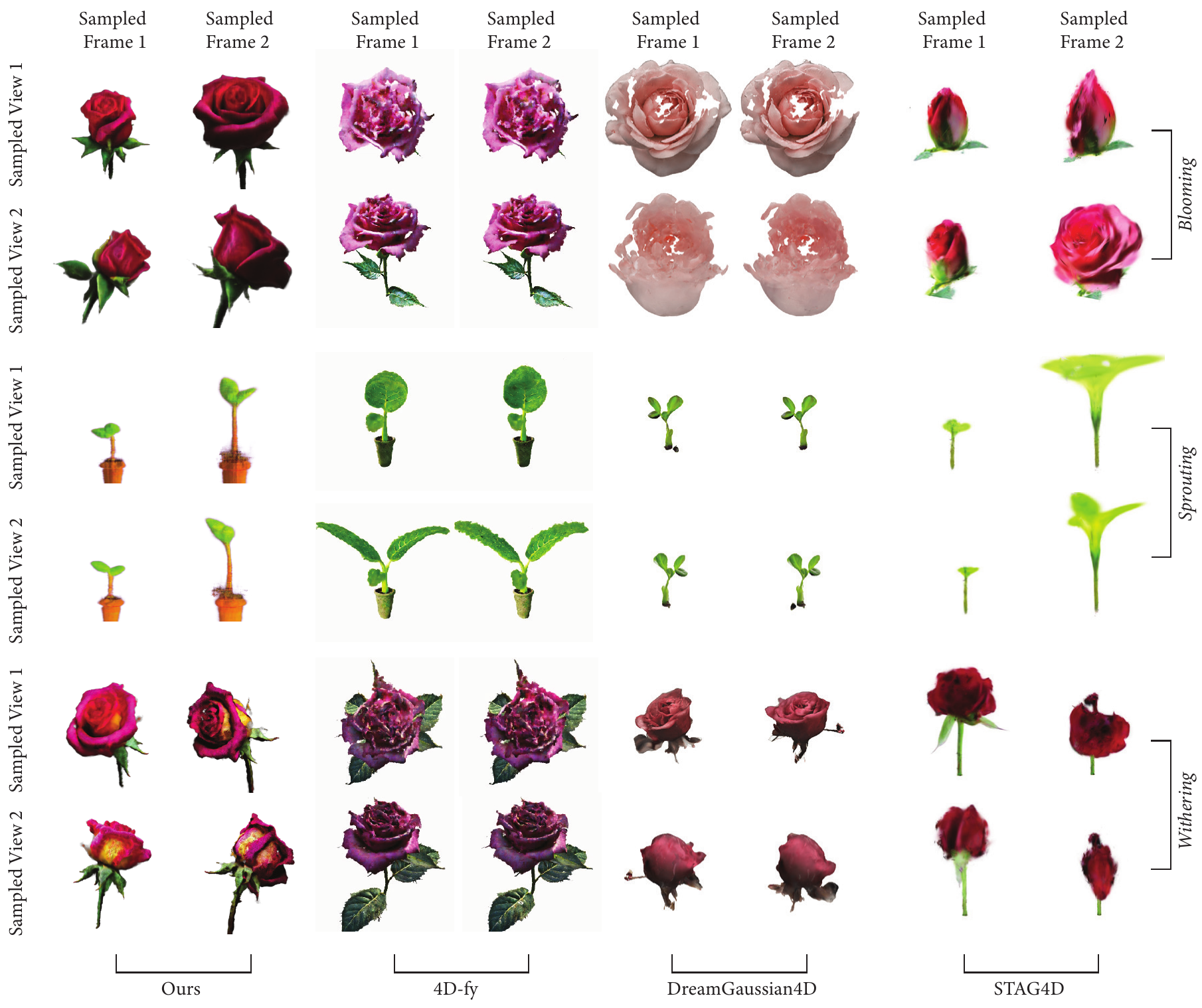}
  \vspace{-0.3cm}
  \caption{\textbf{Comparisons with 4D generation baselines.} We compare with several state-of-the-art 4D generation methods as they are the closest to our setting; however, none of these methods synthesize full dynamic object intrinsics. We observed that 4D-fy~\cite{4dfy} and DreamGaussian4D~\cite{dreamgaussian4d} produce visually appealing static frames, but typically produce very small or unrealistic motions. STAG4D performs poorly in novel view renderings due to the limitation of the 3D reconstruction model.
  }
  \vspace{-15pt}
  \label{fig:comp4d}
\end{figure*}

\section{Experiments}

We test the proposed method on different examples of dynamic natural phenomena, all with a temporal evolving nature of object intrinsics. We also compare our model with state-of-the-art baselines.

\subsection{\Task Generation}

We study 7 different natural phenomena of dynamic objects with significant intrinsics changes: ``flower blooming'', ``flower withering'', ``plant sprouting'', ``candle burning'', ``icecream melting'', ``banana rotting'', and ``bread baking''. \cref{fig:major} shows part of the generation result of 4D object intrinsics from the proposed framework. More results can be found in the supplementary material.

Even for these complex dynamic phenomena with drastic object intrinsics transformation, our generation pipeline produces realistic synthesis results of the intrinsic properties with high temporal consistency in the dynamics motion.

\subsection{Comparison with Prior Works} 

To the best of our knowledge, this work is the first to generate \task for a natural process. Therefore, we choose baselines under the most similar task. 

\myparagraph{4D Generation Methods.}
We compare with recent 4D content generation methods for generic objects, on the 7 different natural phenomena described above. Note that the prior methods designed for this task cannot synthesize albedo or material properties as ours. Therefore, we only evaluate 4D renderings for comparison.

4D-fy \cite{4dfy} synthesizes the 4D content conditioned on text prompts, using guidance from video and image diffusion models and a multi-resolution hash encoding for representation. DreamGaussian4D \cite{dreamgaussian4d} receives an image as input, and synthesizes 4D outputs with a 4D Gaussian Splatting representation.
STAG4D~\cite{stag4d} takes a video input and performs 4D reconstruction on the input video. We use the same reference video as ours to condition STAG4D. The detailed preparation of inputs for the baselines is described in the supplementary material.

Results are shown in \cref{fig:comp4d}.
Our method demonstrates higher rendering quality and better motion consistency compared to the baselines. 4D-fy fails to synthesize dynamic content in accordance with prompted motion concepts.
DreamGaussian4D demonstrates severe temporal and static artifacts.
STAG4D performs 4D reconstruction for a given input video and can fit the input view, but shows noticeable artifacts in novel views. 

We perform a user study comparing the quality of our method with the baselines, where participants are asked to select the most preferred results based on three criteria: concept alignment, motion realism, and overall visual quality. The results of the average rate of preference among 95 participants are reported in \cref{tab:user-study}, showing a clear preference for the proposed methods. Beyond user study, we report CLIP score~\cite{clip} in \cref{tab:user-study} to reflect the objective visual quality and textual alignment.
Additional results and detailed settings are included in the supplementary material.

\begin{figure}[t]
  \centering
  \includegraphics[width=0.7\linewidth]{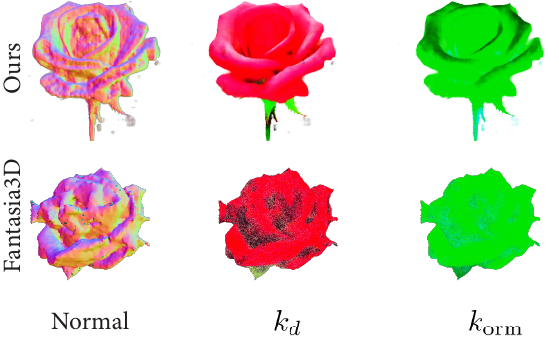}
  \vspace{-0.3cm}
  \caption{\textbf{Comparison with a static object intrinsics generation baseline.} We compare the state-of-the-art Fantasia3D~\cite{fantasia3d}, which synthesizes a ``static'' version of the \task. 
  }
  \vspace{-5pt}
  \label{fig:intrinsics}
\end{figure}

\myparagraph{3D Generation Methods with Material Modeling.} 
Another line of work that tackles part of our task is 3D generation methods with decomposed geometry and appearance. In particular, we compare with Fantasia3D~\cite{fantasia3d}, which receives a text prompt as input (see the supplementary material for details). 
As shown in \cref{fig:intrinsics}, the baseline has a severe Janus problem in geometry and artifacts in texture, while our method synthesizes higher-quality materials.

\begin{figure}[t]
  \centering
  \includegraphics[width=\linewidth]{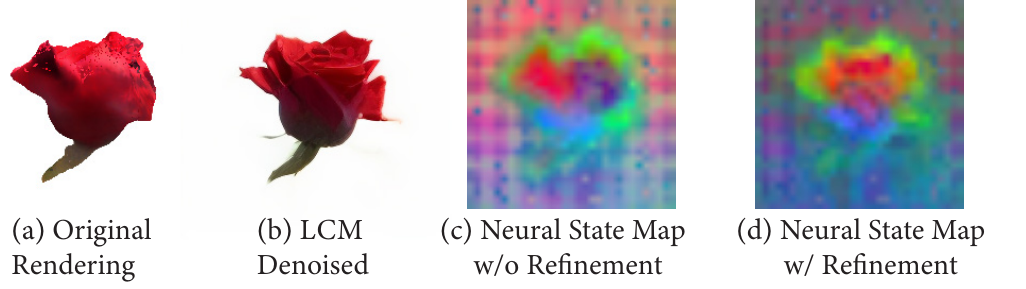}
  \vspace{-18pt}
  \caption{\textbf{Analyzing the influence of using consistency model 
 to perform refinement.} Using the consistency model to do refinement (\cref{ssec:feat_render}) helps mitigate the distribution gap between the raw RGB renderings and natural images and improves the quality of the output \poses~(c-d). 
  }
 \vspace{-5pt}
  \label{fig:abl_lcm}
\end{figure}

\begin{figure}[t]
  \centering
  \includegraphics[width=\linewidth]{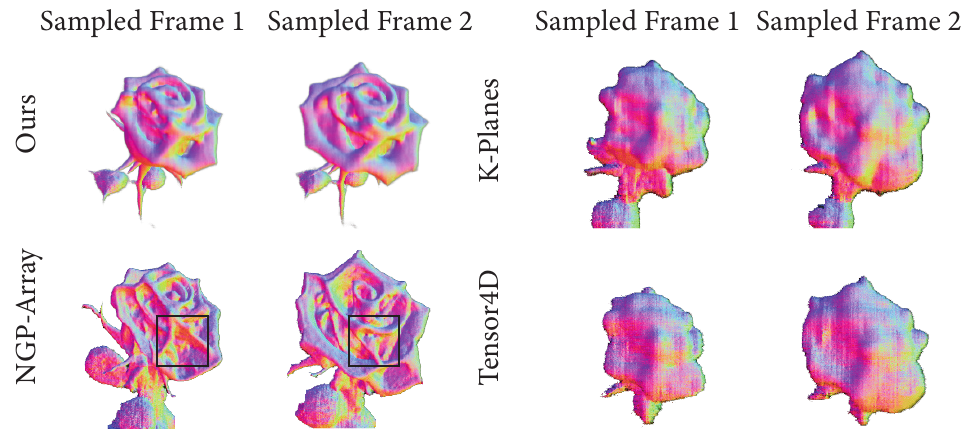}
  \caption{\textbf{Analysis of different 4D representations in guided generation.} We replace the hybrid 4D representation (\cref{ssec:hybrid}) with three alternatives, resulting in either over-smoothing artifacts (K-Planes and Tensor4D) or temporal inconsistency (NGP-Array). }
  \vspace{-15pt}
  \label{fig:abl_4d}
\end{figure}

\subsection{Ablations and Validations}

We ablate several important components in our pipeline.

\label{ssec:ablation_posevol}

\myparagraph{Denoising Renderings with Consistency Model.} We study the effect of the operation to denoise with consistency model (\cref{ssec:feat_render}) by visualizing the denoised RGB rendering and corresponding \poses in \cref{fig:abl_lcm}. The original rendering of the dynamic mesh has a correct temporal state, yet its appearance does not fully align with the natural image distribution DINOv2 encoder $F$ was trained on. By performing denoising, we can obtain a better \pose depicting the temporal state more accurately.

\label{ssec:ablation_synthesis}

\begingroup
\setlength{\columnsep}{8pt}%
\setlength{\intextsep}{1pt}
\begin{wrapfigure}{r}{0.3\textwidth}
  \centering
  \includegraphics[width=0.3\textwidth]{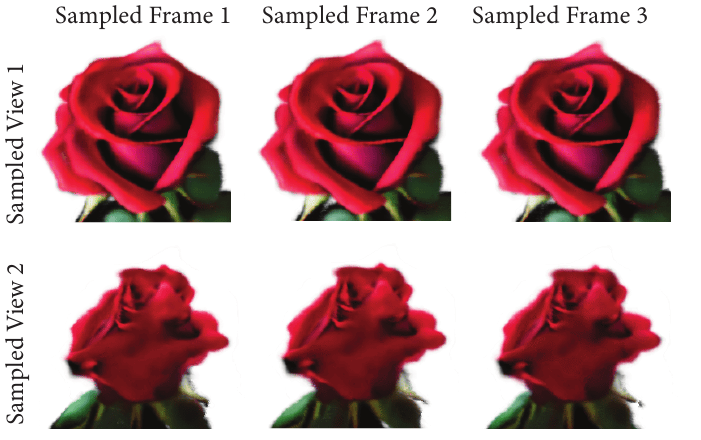}
  \vspace{-20pt}
  \caption{\textbf{Ablation of \posevol guidance.}}
  \label{fig:abl_pose_field}
\end{wrapfigure}
\myparagraph{Generation without Guidance.} We show that using guidance during generation is necessary. Removing the Neural Template guidance in the pipeline leads to results in \cref{fig:abl_pose_field}, with visual artifacts of 3D inconsistency across views and the tiny motion across frames, suggesting the importance of the guidance.

\endgroup

\myparagraph{Architecture of Hybrid 4D Representations.} We further explore different variants of 4D representation used in the generation pipeline. We experiment with other possible representations, including pure K-Planes \cite{kplanes2023}, Tensor4D \cite{shao2023tensor4d}, and an array of NGP encodings \cite{muller2022instant}. The results can be found at \cref{fig:abl_4d}. K-Planes and Tensor4D produce smooth outputs but struggle to represent the fine details. An array of NGP encodings is expressive enough but manifests poor temporal consistency, while the proposed representation achieves the best overall quality.

\section{Conclusion}

We have presented a novel and important task of generating 4D \task only from the guidance of 2D foundation models, proposed a new representation of \posevol, and explored its use in generating \task for different phenomena. %

\paragraph{Acknowledgments.}
The paper's title is inspired by the artwork, \emph{Nacer y Morir de una Rosa}, by Colombian artist Rosa Navarro, recently on exhibit at the Orange County Museum of Art in 2023. We thank Haian Jin, Zizhang Li, and Ruocheng Wang for insightful discussions and feedback on the paper. This work is in part supported by NSF RI \#2211258, \#2338203, ONR MURI N00014-22-1-2740, and Samsung. YZ is in part supported by the Stanford Interdisciplinary Graduate Fellowship.

{
    \small
    \bibliographystyle{ieeenat_fullname}
    \bibliography{ref}
}

\end{document}